\documentclass{article}

\usepackage[preprint]{neurips_2026}
\makeatletter
\renewcommand{\@noticestring}{Accepted at the DiffuLM Workshop at NeurIPS 2026.}
\makeatother

\usepackage[utf8]{inputenc}
\usepackage[T1]{fontenc}
\usepackage{amsmath}
\usepackage{amssymb}
\usepackage{booktabs}
\usepackage{microtype}
\usepackage{xcolor}
\usepackage{hyperref}
\hypersetup{
  hidelinks,
  pdftitle={Hidden Messages Between Diffusion Language Model Agents: Causal Evidence for Latent Communication},
  pdfauthor={Alan-Barsag Gazzaev, Alexey Gavrilov, Sergey Muravyov, Ilya Makarov}
}
\usepackage{url}
\usepackage{tikz}
\usetikzlibrary{arrows.meta,calc,fit,positioning}

\definecolor{agentblue}{HTML}{DDEBFA}
\definecolor{agentgreen}{HTML}{DFF3E4}
\definecolor{latentviolet}{HTML}{E8DDF5}
\definecolor{causalorange}{HTML}{FCE7CF}
\definecolor{inkblue}{HTML}{244A73}
\definecolor{inkgreen}{HTML}{27613A}
\definecolor{inkviolet}{HTML}{5C3A7A}
\definecolor{inkorange}{HTML}{8B4C13}

\newcommand{\method}{\textsc{Hidden Messages}}
\newcommand{\msg}{\mathbf{z}}

\newcommand{\state}{\mathbf{h}}
\newcommand{\receive}{\mathbf{r}}

\newcommand{\TinyNoMessage}{8.6\%}
\newcommand{\TinyMatched}{72.9\%}
\newcommand{\TinyDeranged}{9.3\%}

\newcommand{\TinyContentGain}{+63.6}
\newcommand{\TinyContentCI}{[$+61.5$, $+65.8$]}
\newcommand{\TinyWrongFactTarget}{72.9\%}
\newcommand{\TinyWrongFactFlip}{89.4\%}
\newcommand{\SyntheticNoMessage}{9.8\%}
\newcommand{\SyntheticMatched}{100.0\%}
\newcommand{\SyntheticDeranged}{0.0\%}
\newcommand{\SyntheticLate}{10.8\%}
\newcommand{\SyntheticContentGain}{+100.0}

\newcommand{\SyntheticWrongFactTarget}{100.0\%}
\newcommand{\QASCNoMessage}{72.6\%}
\newcommand{\QASCRecurrent}{80.8\%}
\newcommand{\QASCDeranged}{46.8\%}
\newcommand{\QASCLate}{87.8\%}
\newcommand{\QASCPresenceGain}{+8.2}
\newcommand{\QASCPresenceCI}{[$+5.7$, $+10.7$]}
\newcommand{\QASCContentGain}{+34.0}
\newcommand{\QASCContentCI}{[$+30.7$, $+37.2$]}
\newcommand{\QASCLateGain}{+7.0}
\newcommand{\QASCLateCI}{[$+4.4$, $+9.6$]}

\title{Hidden Messages Between Diffusion Language Model Agents:\\
Causal Evidence for Latent Communication}

\author{%
  Alan-Barsag Gazzaev \\
  ITMO University \\
  \texttt{avgazzaev@itmo.ru}
  \And
  Alexey Gavrilov \\
  ITMO University \\
  \texttt{aogavrilov@itmo.ru}
  \AND
  Sergey Muravyov \\
  ITMO University \\
  \texttt{smuravyov@itmo.ru}
  \And
  Ilya Makarov \\
  HSE University \\
  \texttt{iamakarov@hse.ru}
}

\begin{document}

\maketitle

\begin{abstract}
Diffusion language models form answers through a sequence of revisable continuous states, but
multi-agent systems usually discard those states and make agents communicate only after decoding
text.  We ask whether one independent diffusion trajectory can pass a compact hidden message into
another and, crucially, whether the receiver uses what that message means.  In \method, a sender
compresses its denoising state into four signed dense vectors; the receiver consumes them while
retaining its own context, canvas, and trajectory.  Removing a message tests whether the channel
matters; replacing it with a message from another example tests whether its content matters.
Across controlled tasks, derangement removes a \TinyContentGain{}-point advantage on Tiny-A2D and
a \SyntheticContentGain{}-point advantage on Dream-7B, while targeted substitutions steer the
receiver toward the substituted fact.  On QASC with the two supporting facts split across agents,
matched communication reaches \QASCRecurrent{}, but a wrong-example message lowers accuracy to
\QASCDeranged{}, a content effect of \QASCContentGain{} points.  The surprising result is temporal:
one late transfer reaches \QASCLate{}, exceeding four recurrent exchanges by \QASCLateGain{}
points.  These experiments provide causal evidence for continuous hidden communication between
diffusion-LM agents and show
that \emph{when} information is transferred can matter more than how many times agents exchange it.
\end{abstract}

\section{Introduction}

Masked diffusion language models generate by repeatedly revising a partially masked sequence
rather than committing to a left-to-right token stream \citep{li2022diffusionlm,sahoo2024mdlm,
nie2025llada,ye2025dream,zhou2026dllm}.  Between the initial mask and the final answer lies a rich
sequence of continuous states.  A single model uses those states internally, then throws them away.
A group of agents could instead treat them as a communication surface.

Imagine two agents looking at different pieces of the same problem.  One knows which fact is
needed; the other holds a table containing that fact.  A text-only protocol makes the first agent
finish a sentence before the second can use the request.  A hidden-state protocol can hand over a
compact continuous representation while both answers are still being revised.  This is the object
we study: messages between \emph{independent diffusion-LM agents}, with no shared canvas, context,
token prediction, or trajectory.  Recent latent multi-agent systems relay hidden states or
key--value caches between autoregressive models \citep{du2026interlat,zou2026latentmas}, while
other work preserves state within one diffusion model or transfers partially denoised sequences
\citep{xia2026metastate,rozonoyer2026relay,yun2026tie}.  Here the boundary is agent-to-agent and
continuous, but each denoising process remains its own.  We train a task-specific communication
adapter with supervision around frozen backbones: the question is whether this learned channel
carries causally useful content, not whether a communication protocol emerges without supervision.

An accuracy increase alone, however, is not evidence that a hidden message conveys the sender's
example-specific content.  The receiver could benefit from a generic activation, extra compute,
or a distributional cue.  Causal audits of latent channels have made this distinction explicit
for autoregressive agents \citep{zhang2026causalaudit,cheng2026kvaudit}.  We bring this standard to
diffusion-LM collaboration: the message entering the receiver is replaced while everything else
is held fixed.  A matched message is compared both with no message and with a \emph{deranged}
message produced for another example.  The first contrast asks whether the channel matters; the
second asks whether the receiver uses the right content.

We follow the idea from its smallest causal test to a natural reasoning task.  Tiny-A2D asks
whether a compact diffusion backbone can learn an example-specific channel at all.  The Dream-7B
dependency-spanning task then asks whether the same causal signature survives scale and a dependency that
requires information to travel out and back.  Finally, distributed QASC asks whether the channel
changes answers when scientific evidence is split across agents.  This progression exposes a
result that a simple
``more interaction is better'' comparison would miss: repeated exchange is indispensable for the
out-and-back dependency, yet one late transfer beats four recurrent exchanges on QASC.  The
relevant design variable is not raw round count but the match between a task's dependency
structure and the moment at which a hidden state becomes useful.

Our contributions are threefold:

\begin{itemize}
  \item We introduce a supervised signed dense message channel for diffusion-LM agents and instantiate it on
  Tiny-A2D and Dream-7B while preserving independent contexts, canvases, random-number streams,
  and denoising trajectories.
  \item We establish content-sensitive communication through receiver-side interventions:
  removing, deranging, zeroing, randomizing, or self-substituting messages while retaining a fixed
  designated receiver.
  \item We identify a temporal principle across three empirical settings: recurrent communication
  helps when dependencies are sequential, whereas late fusion can be both simpler and more
  accurate when private evidence is independently computable.
\end{itemize}

\noindent\textbf{Code and artifacts:}\\
{\small\url{https://github.com/alanrbtx/hidden-messages-diffusion-agents}}

\section{Related work}

\paragraph{Latent multi-agent communication.}
Interlat learns compressed hidden-state communication between language-model agents
\citep{du2026interlat}; LatentMAS relays latent thoughts through shared working memory
\citep{zou2026latentmas}; and Latent-DARM bridges diffusion planners and autoregressive executors
\citep{berrayana2026latentdarm}.  These works establish that text is not the only viable agent
interface.  Our focus is the denoising-time interface between diffusion-LM agents and the causal
identity of a compact message, with independent rather than shared canvases.  This distinction
turns communication into a testable object: the sender-side message can be replaced without
replacing the receiver's context or trajectory.

\paragraph{State across diffusion steps.}
MetaState and Relay preserve continuous computation across successive passes of one diffusion
model \citep{xia2026metastate,rozonoyer2026relay}.  TIE transfers promising partially denoised
sequences across models \citep{yun2026tie}.  \method{} instead transports a compressed hidden
message between separate agents; the receiving canvas itself is never replaced.  This distinction
is important for attributing an effect to communication rather than trajectory selection.

\paragraph{Causal audits.}
Recent audits show that message presence, example-specific content, and sender identity are
separate properties of latent channels \citep{zhang2026causalaudit,cheng2026kvaudit}.  We adopt
matched-versus-deranged replacement as a necessary content test and extend the intervention axis
to delivery time.  The resulting question is not only whether a vector matters, but whether the
right vector arrives when the receiving computation can use it.

\section{Hidden communication between diffusion agents}

\subsection{Independent trajectories and a continuous interface}

Consider agents $i\in\{0,\ldots,N-1\}$.  Agent $i$ has private context $c_i$, a masked canvas
$x_i^t$, and its own denoising trajectory.  At an exchange point $t$, an encoder reads the
task-relevant part of hidden state $\state_i^t$ and produces a compact message
\begin{equation}
  \msg_i^t = E_\theta(\state_i^t) \in \mathbb{R}^{K\times d},
  \qquad K=4,\ d=128.
  \label{eq:encode}
\end{equation}
The transmitted object is a signed BF16 tensor, not token IDs, class probabilities, logits, or a
partially decoded sequence.  Incoming messages from other agents are averaged and decoded into a
receiver-side conditioning state,
\begin{equation}
  \bar{\msg}_i^t = \frac{1}{N-1}\sum_{j\neq i}\msg_j^t,
  \qquad
  \receive_i^t = D_\phi(\bar{\msg}_i^t).
  \label{eq:project}
\end{equation}
On Tiny-A2D, $E_\theta$ attends over the sender token states and $D_\phi$ fuses the decoded message
into the receiver's answer state through a gated residual.  On Dream-7B, the encoder reads the
answer-position state and the decoder produces $K$ virtual prefix embeddings, RMS-matched to the
backbone input embeddings and prepended to the receiver's next pass.  In recurrent exchange the
prefix persists until it is replaced.  No agent sees another agent's text context or canvas.

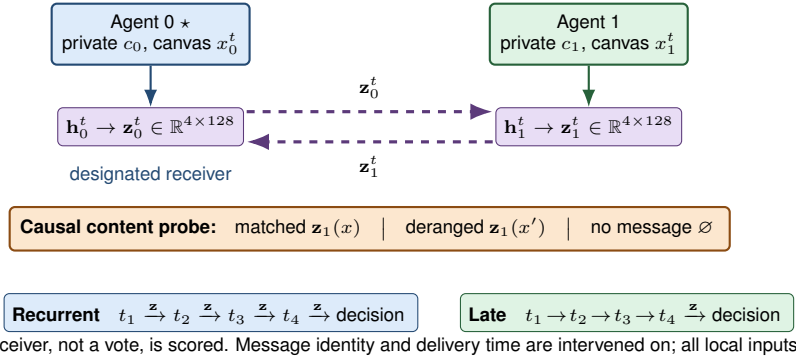
\begin{figure}[t]
\centering
\begin{tikzpicture}[
  >=Latex,
  font=\sffamily\scriptsize,
  box/.style={rounded corners=2pt, draw, thick, align=center, inner sep=4pt},
  smallbox/.style={rounded corners=2pt, draw, align=center, inner sep=3pt},
  flow/.style={->, thick},
  msg/.style={->, very thick, dashed, draw=inkviolet, shorten <=1pt, shorten >=1pt},
  probe/.style={->, thick, draw=inkorange},
]
  \node[box, fill=agentblue, draw=inkblue, minimum width=2.55cm] (a0)
    {Agent 0 $\star$\\private $c_0$, canvas $x_0^t$};
  \node[box, fill=agentgreen, draw=inkgreen, minimum width=2.55cm, right=3.15cm of a0] (a1)
    {Agent 1\\private $c_1$, canvas $x_1^t$};
  \node[smallbox, fill=latentviolet, draw=inkviolet, below=0.55cm of a0] (z0)
    {$\state_0^t \rightarrow \msg_0^t\in\mathbb{R}^{4\times128}$};
  \node[smallbox, fill=latentviolet, draw=inkviolet, below=0.55cm of a1] (z1)
    {$\state_1^t \rightarrow \msg_1^t\in\mathbb{R}^{4\times128}$};
  \draw[flow, draw=inkblue] (a0) -- (z0);
  \draw[flow, draw=inkgreen] (a1) -- (z1);
  \draw[msg]
    ([yshift=2.0mm]z0.east) --
    node[midway, above=1.2pt] {$\msg_0^t$}
    ([yshift=2.0mm]z1.west);
  \draw[msg]
    ([yshift=-2.0mm]z1.west) --
    node[midway, below=1.2pt] {$\msg_1^t$}
    ([yshift=-2.0mm]z0.east);
  \node[below=0.15cm of z0, text=inkblue] {designated receiver};

  \node[box, fill=causalorange, draw=inkorange, below=1.05cm of $(z0)!0.5!(z1)$,
        minimum width=8.25cm] (audit) {
    \textbf{Causal content probe:}\quad
    matched $\msg_1(x)$\quad $\big|$\quad
    deranged $\msg_1(x')$\quad $\big|$\quad
    no message $\varnothing$
  };
  \node[smallbox, fill=agentblue, draw=inkblue, below=0.55cm of audit, xshift=-2.1cm,
        minimum width=3.7cm] (recur) {
    \textbf{Recurrent}\quad
    $t_1\xrightarrow{\msg}t_2\xrightarrow{\msg}t_3\xrightarrow{\msg}t_4\xrightarrow{\msg}$ decision
  };
  \node[smallbox, fill=agentgreen, draw=inkgreen, right=0.55cm of recur,
        minimum width=3.7cm] (late) {
    \textbf{Late}\quad
    $t_1\!\rightarrow\!t_2\!\rightarrow\!t_3\!\rightarrow\!t_4\xrightarrow{\msg}$ decision
  };
  \node[below=0.2cm of $(recur)!0.5!(late)$, align=center] {
    The receiver, not a vote, is scored.  Message identity and delivery time are intervened on;
    all local inputs stay fixed.
  };
\end{tikzpicture}
\caption{\method{} exchanges compact continuous states while each diffusion-LM agent keeps an
independent canvas and trajectory (top).  The receiver consumes the four slots through residual
fusion on Tiny-A2D or a virtual prefix on Dream-7B.  Matched-versus-deranged replacement isolates
example-specific content; recurrent-versus-late delivery isolates timing (bottom).  The star marks
the predesignated receiver.}
\label{fig:method}
\end{figure}

\subsection{Supervised task-specific channel adaptation}

The channel is a lightweight, task-specific adapter comprising a message encoder and a receiver-side
fusion or projection module.  The backbone and language-model head are frozen in every reported
channel experiment; supervision trains the communication interface rather than the diffusion models.  For
Tiny-A2D, the compressor and residual fuser are learned jointly: union-context hidden states act as
a teacher, while paired matched-versus-deranged supervision makes messages discriminate between
counterfactual sender facts.  For Dream-7B \citep{ye2025dream}, an auxiliary classifier first
encourages the message to preserve the requested slot, returned code, or QASC answer label; a
variance term prevents collapse.  We then freeze the encoder and align the receiver-side projector
to short answer-oriented prefix embeddings using cosine and relative mean-squared error.
Supervision shapes the message geometry, but neither the label, teacher context, nor any text
message is available to the receiver at inference time.  The reported settings use task-specific
channel training, not a single universal adapter transferred across all tasks.  Exact optimization settings are given in
Appendix~\ref{app:implementation}.

\subsection{Causal and temporal interventions}

Let $Y(x,\msg)$ indicate whether the designated receiver answers example $x$ correctly under
message input $\msg$.  We evaluate three core interventions:
\begin{align}
  \Delta_{\mathrm{presence}} &= \mathbb{E}[Y(x,\msg(x))-Y(x,\varnothing)],\\
  \Delta_{\mathrm{content}}  &= \mathbb{E}[Y(x,\msg(x))-Y(x,\msg(\pi(x)))],
  \label{eq:contrasts}
\end{align}
where $\pi$ is a deterministic permutation to a different-target example whenever possible.
Because the receiver input, local trajectory, and evaluation item are unchanged, the second
contrast attributes the difference to message identity rather than mere channel presence.  We
also compare recurrent delivery with \emph{late-only} delivery: agents first process private
evidence independently, then exchange once immediately before the receiver decision.  Zero,
moment-matched random, and self-message replacements serve as additional checks (Appendix~\ref{app:controls}).

\section{Experimental setup}

\subsection{Two backbones, three views of the channel}

\paragraph{HM-Synth Chain on Tiny-A2D.}
We begin with the minimal causal question: can information cross the hidden boundary at all?
HM-Synth generates chains of two to four lookup facts and splits them across two private contexts.
Each counterfactual pair keeps the designated receiver's context fixed but changes a sender-side
fact and therefore the correct answer.  The frozen Tiny-A2D backbone sees 10,000 training examples
and no union context at inference; only a 6.1 MiB communication module is learned.  We evaluate
1,000 counterfactual pairs (2,000 examples), making both message derangement and targeted fact
replacement unambiguous.
This benchmark acts as a microscope for the channel: when the receiver input is fixed and only the
sender fact changes, a corresponding answer change has only one route across the agent boundary.

\paragraph{Dream-7B dependency spanning.}
Agent 0 knows which of ten slots is requested but does not have the code table.  Agent 1 has the
complete, randomly ordered table but is not told which slot is requested.  Solving the task
requires a two-stage out-and-back exchange: agent 0 communicates the requested slot, agent 1 uses that message to select
the relevant code, and agent 1 communicates the code back before agent 0 commits its answer.  We
evaluate 512 counterfactual pairs (1,024 examples) with disjoint private contexts.  In addition to
example derangement, a wrong-fact intervention swaps a table entry while preserving agent 0's
context; this lets us test whether the final prediction follows the substituted content.
The first transfer is therefore a query and the second a reply; a single late handoff cannot
recover the intermediate computation that never took place.

\paragraph{Distributed QASC.}
QASC questions are annotated with two facts whose composition supports an eight-way answer
\citep{khot2020qasc}.  We give one gold fact to each of two agents, add four disjoint
leakage-filtered distractors per agent, and keep the question and options common.  Neither agent
sees the other's evidence.  The channel is learned only from the 7,134 derived training examples
and evaluated on all 1,000 derived development examples.  Agent 0 is fixed as receiver in every
condition; there is no voting or logit averaging.
This setting replaces synthetic codes with ordinary scientific statements while preserving the
central information constraint: the scored receiver must combine evidence that it never observes
in one text context.

\paragraph{Inference and uncertainty.}
For QASC, recurrent communication consists of four synchronous exchanges followed by one receiver
decision; late-only communication consists of one independent pass per agent, one exchange, and
the same receiver decision.  We reuse identical item order and receiver inputs across conditions.
HM-Synth and QASC accuracy contrasts use 5,000 and 10,000 paired bootstrap resamples over examples,
respectively, with percentile 95\% intervals.  The Dream-7B dependency-spanning row in
Table~\ref{tab:main-results} reports the aggregate over locked evaluations; its item-level causal
contrasts have the same direction throughout.

\section{Results}

\begin{table}[t]
\centering
\caption{Three views of the hidden channel.  Accuracies are percentages and contrasts are
percentage points.  Tiny-A2D establishes content transfer on HM-Synth; the Dream-7B dependency-spanning task
tests a sequential dependency, and distributed QASC tests split natural evidence.  The schedule
reversal in the last column is the central timing result: late fusion fails on the Dream-7B
dependency-spanning task but wins on QASC.}
\label{tab:main-results}
\small
\setlength{\tabcolsep}{2.7pt}
\begin{tabular}{lrrrrrr}
\toprule
Setting & No msg. & Matched & Deranged & Late & Match.$-$der. & Late$-$match. \\
\midrule
HM-Synth (Tiny-A2D) & 8.6\% &
72.9\% & 9.3\% &
\multicolumn{1}{c}{--} & $+63.6$ &
\multicolumn{1}{c}{--} \\
Dependency spanning (Dream-7B) & 9.8\% &
100.0\% & 0.0\% &
10.8\% &
$+100.0$ &
$-89.2$ \\
QASC (Dream-7B) & 72.6\% & 80.8\% &
46.8\% & 87.8\% &
$+34.0$ & $+7.0$ \\
\bottomrule
\end{tabular}
\vspace{1pt}

\footnotesize Tiny-A2D content CI: [$+61.5$, $+65.8$];
QASC content CI: [$+30.7$, $+37.2$];
QASC late$-$matched CI: [$+4.4$, $+9.6$].
\end{table}

\subsection{The receiver follows the message, not merely the channel}

Tiny-A2D supplies the first causal trace.  A matched message raises HM-Synth accuracy from
\TinyNoMessage{} to \TinyMatched{}; a message from a different-target example returns performance
to \TinyDeranged{}.  The matched-minus-deranged effect is \TinyContentGain{} points with 95\%
paired bootstrap CI \TinyContentCI, while random, zero, and self messages all remain near chance.
More revealingly, when we replace the sender's fact, the receiver changes its answer in
\TinyWrongFactFlip{} of examples and selects the substituted target in \TinyWrongFactTarget{}.
The vector is therefore carrying a decision-relevant fact, not merely opening a learned gate.

The Dream-7B dependency-spanning task reproduces the same signature in a harder out-and-back exchange.  Matched two-stage
communication reaches \SyntheticMatched{}, compared with \SyntheticNoMessage{} without a message
and \SyntheticDeranged{} under derangement.  Swapping the table fact steers the receiver to the
replacement target in \SyntheticWrongFactTarget{} of examples.  Scaling the backbone does not
wash out the causal dependence on message identity; it sharpens it.

The same causal signature appears on natural questions.  QASC accuracy rises from
\QASCNoMessage{} without communication to \QASCRecurrent{} with matched recurrent messages.  The
\QASCPresenceGain{}-point paired gain has interval \QASCPresenceCI.  More decisively, replacing
each message with one produced for a different-target question reduces accuracy to
\QASCDeranged{}: the matched-minus-deranged content effect is \QASCContentGain{} points with
interval \QASCContentCI.  Wrong content is substantially worse than silence.  A channel whose
benefit came only from additional computation or a constant learned prefix would not be sensitive
to which question produced the message, much less be harmed by receiving the wrong one.

\subsection{The right moment beats more rounds}

The direction of the schedule effect reverses with the information structure.  The Dream-7B
dependency-spanning task needs messages at two different points: agent 1 cannot select the requested code before
receiving the slot identity.  Recurrent exchange reaches \SyntheticMatched{}, but one late message
reaches only \SyntheticLate{}, near the no-communication level.  The first message is useful not
because it contains the final answer, but because it changes what the other agent computes next.

QASC has a different structure.  Each agent can interpret its own supporting fact before any
exchange.  Here, a single late transfer reaches \QASCLate{}, outperforming the four-round recurrent
schedule by \QASCLateGain{} points (\QASCLateCI).  The late schedule transmits only once and uses
three agent forward passes, compared with four transmissions and nine forward passes for the
recurrent schedule, yet it is more accurate.  Repeatedly injecting provisional states appears to
interfere with independent evidence formation; delaying fusion lets the sender first form a more
decision-ready representation.

The two regimes suggest a temporal design rule.  If one agent's next computation is conditional on
another's partial result, the message must arrive early enough to redirect that computation.  If
private evidence can mature independently, early repeated fusion can be counterproductive and the
best handoff is close to the decision.  Round count is therefore a poor proxy for collaboration
depth: the useful quantity is whether the exchange point aligns with the task's latent dependency
graph.

\section{Discussion}

\paragraph{What travels through the channel.}
Matched accuracy alone would leave the mechanism ambiguous: a learned prefix might simply make the
receiver more confident or place it in a generally favorable activation regime.  The replacement
experiments tell a more specific story.  Messages produced for the wrong example hurt, while
targeted fact substitutions redirect the receiver toward the substituted answer.  We therefore do
not need to translate the vectors into sentences to test whether they carry decision-relevant
content; changing their origin while holding the receiver fixed changes the decision in a
predictable direction.  This supports hidden communication, not a claim of a human-readable or
emergent language.

\paragraph{Communication has a clock.}
The schedule reversal shows why an agent protocol cannot be summarized by its round count.  In the
Dream-7B dependency-spanning task, the first message determines what the other agent should compute, so waiting
until the end removes the opportunity to perform that computation.  In QASC, each agent can first
form a useful representation of its own evidence, and early repeated fusion can disturb that
process.  Both settings use the same Dream-7B backbone, so the reversal is not a trivial consequence
of comparing different model sizes; it is consistent with the dependency structure and the point
at which a message becomes actionable.  The broader design opportunity is an adaptive protocol
that exchanges when a latent dependency is ready, rather than on a fixed clock.

\section{Limitations and future work}

The controlled evidence spans Tiny-A2D and Dream-7B, but natural-question evidence currently comes
from Dream-7B on QASC; larger teams, other diffusion backbones, and long-form generation may produce
different channel geometries.  The channels are supervised, task-specific adapters, so the results
establish learned hidden communication but not an emergent language or a task-independent protocol.
Initial zero-shot reuse across heterogeneous tasks did not yield a detectable benefit.  Applying the
channel to a new task may therefore require new supervision and adapter retraining or explicit
cross-domain alignment; successful zero-shot transfer is not established by these experiments.
Matched text-channel and wall-clock comparisons are also future work.  Finally, a latent channel is
not a privacy mechanism: dense states may encode sensitive information even when they are not
human-readable.  A particularly consequential next setting is multi-agent code generation with
deliberately partitioned context: different agents may see the specification, relevant source
files, tests, or execution feedback.  Continuous hidden communication could carry task-specific
intermediate state across these boundaries without requiring every exchange to be verbalized, while
matched-versus-deranged interventions would test whether successful code changes causally depend on
the transmitted content.  This setting also sharpens the timing question: a message should arrive
when a cross-agent dependency becomes actionable, rather than at every decoding round.  We therefore
view context-partitioned software engineering, together with a content-dependent message policy, as
the primary direction for scaling the method.

\section{Conclusion}

With a supervised task-specific adapter, diffusion-LM agents can exchange useful information without
verbalizing it.  From Tiny-A2D
HM-Synth to the Dream-7B dependency-spanning task and distributed QASC, removing, deranging, or surgically replacing
the hidden message changes a fixed receiver in the predicted direction.  That sequence of
interventions is the causal evidence: example-specific content crosses the latent boundary.  The
central result is not that more messages are always better.  Two-stage exchange is essential when
one agent's computation depends on another's earlier message, while a single late transfer is
superior when private evidence can mature independently.  Hidden communication is therefore a
temporal coordination problem: the moment of transfer can matter more than the number of rounds.

\label{maintextend}

\clearpage
\begin{ack}
This paper was carried out as part of ITMO University project No.~666001
``Laboratory of Multi-Agent Modeling and Adaptive Intelligence''.

The authors declare no competing interests.
\end{ack}

\bibliographystyle{plainnat}
\bibliography{references}

\appendix

\section{Implementation and evaluation details}
\label{app:implementation}

We use the Tiny-A2D checkpoint \texttt{dllm-hub/Qwen3-0.6B-diffusion-mdlm-v0.1}
\citep{zhou2026dllm} and Dream-v0-Instruct-7B, both with BF16 weights.  Dream uses SDPA attention,
and both model checkpoints are released under Apache-2.0.  QASC is used under CC-BY-4.0.  Exact
pinned model and dataset revisions are recorded with the released configurations rather than in
the paper prose.  The derived QASC split contains 7,134 training, 1,000 development, and 926 held-out
examples; this paper reports the full development set and does not select examples by method
output.

HM-Synth contains 5,000 training, 500 validation, and 1,000 test counterfactual pairs, with disjoint
generator ranges and two to four hops.  The Tiny-A2D channel has four 128-dimensional slots and
width 256.  Its 960-step AdamW run uses batch size 64, learning rate $3\times10^{-4}$, weight decay
$10^{-2}$, union-teacher weight 0.5, and paired-separation weight 0.2 on one quarter of each batch.
Only split private contexts are available at evaluation.

The Dream encoder is LayerNorm--Linear--SiLU--Linear with per-slot LayerNorm; its
Linear--SiLU--Linear prefix projector is RMS-calibrated to input embeddings.  Representation
learning combines answer-class cross-entropy with a feature-variance penalty (weight 0.1, target
standard deviation 0.2); prefix alignment uses cosine distance plus 0.25 relative MSE.  Both stages
use AdamW for 1,000 steps, batch size 128, learning rate $10^{-3}$, weight decay $10^{-4}$,
gradient clipping at 1.0, and a 5\% warmup.

\paragraph{Code and artifacts.}
The implementation, exact configurations and revisions, learned channel weights, per-example
predictions, and table-generation code are available at
\url{https://github.com/alanrbtx/hidden-messages-diffusion-agents}.

Tiny-A2D used an NVIDIA Quadro RTX 8000 with CUDA 13.0 and PyTorch 2.12 for 1,293.3 seconds
(0.36 GPU-hours).  Dream used NVIDIA H100 80GB GPUs with the same stack: QASC evaluation took
754.6 seconds (0.21 GPU-hours), and controlled channel training plus evaluation took 8,988.7
seconds (2.50 GPU-hours).  These totals exclude download, artifact verification, and other runs.

\section{Intervention details}
\label{app:controls}

\paragraph{Deranged message.}
The sender message is permuted across evaluation rows while the receiver's prompt, private context,
and target stay fixed.  For QASC, the permutation is constrained to a different answer label.  For
the controlled task, counterfactual pairs and different-slot batches ensure that both fact and slot
replacements are defined.

\paragraph{Zero, random, and self controls.}
The zero condition replaces every message element with zero.  The random condition samples a
tensor with the empirical first and second moments of the matched messages at that round.  The
self condition injects an agent's own message through the same projection path.  These conditions
help separate task content from vector presence, magnitude, and receiver-side activation.

\paragraph{Integrity.}
Every agent receives a separate token row and maintains its own diffusion state.  Communication-off
predictions are exactly identical to the unmodified backbone path.  The primary output is always
agent 0; no vote, canvas replacement, logit averaging, or union-context oracle is used.  Paired
bootstrap intervals resample evaluation rows with replacement and recompute the accuracy
difference for 10,000 replicates.

\end{document}